\documentclass[sigconf]{acmart}
\AtBeginDocument{%
 }

\copyrightyear{2025}
\acmYear{2025}
\setcopyright{rightsretained}
\acmConference[MM '25]{Proceedings of the 33rd ACM International Conference on Multimedia}{October 27-31, 2025}{Dublin, Ireland}
\acmBooktitle{Proceedings of the 33rd ACM International Conference on Multimedia (MM '25), October 27-31, 2025, Dublin, Ireland}
\acmDOI{10.1145/3746027.3754911}
\acmISBN{979-8-4007-0686-8/24/10}

\makeatletter
\gdef\@copyrightpermission{
  \begin{minipage}{0.3\columnwidth}
   \href{https://creativecommons.org/licenses/by/4.0/}{\includegraphics[width=0.90\textwidth]{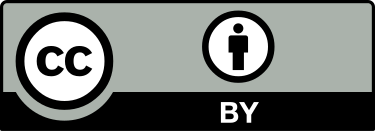}}
  \end{minipage}\hfill
  \begin{minipage}{0.7\columnwidth}
   \href{https://creativecommons.org/licenses/by/4.0/}{This work is licensed under a Creative Commons Attribution International 4.0 License.}
  \end{minipage}
  \vspace{5pt}
}
\makeatother

\usepackage{multirow}
\usepackage{pifont}
\begin{document}

\title{Interact-Custom: Customized Human Object Interaction\\Image Generation}
\begin{abstract}
Compositional Customized Image Generation aims to customize multiple target concepts within generation content, which has gained attention for its wild application. Though a great success, existing approaches mainly concentrate on the target entity's appearance preservation, while neglecting the fine-grained interaction control among target entities. 
To enable the model of such interaction control capability, we focus on human object interaction scenario and propose the task of Customized Human Object Interaction Image Generation (CHOI), which simultaneously requires identity preservation for target human object and the interaction semantic control between them.
We attribute two primary challenges of CHOI as follows:
(1) the simultaneous identity preservation and interaction control demands require the model to decompose the human object into self-contained identity features and pose-oriented interaction features, while the current HOI image datasets fail to provide ideal samples for such feature-decomposed learning.
(2) inappropriate spatial configuration between human and object may lead to the lack of desired interaction semantics, as it may provide wrong hints on the human object body parts crucial for interaction semantic expression. 
To tackle the above issues, we first collect and process a large-scale dataset, where each sample encompasses the same pair of human object involving different interactive poses.  Such data is tailored for CHOI training, from where the model can learn how to decompose identity features and interaction features for target human and object. Then to provide appropriate spatial configuration for interaction semantic expression, we design a two-stage model Interact-Custom, which 
firstly explicitly model the spatial configuration by generating a foreground mask depicting the interaction behavior, then under the guidance of this mask, we generate the target human object interacting while preserving their identities features.
Furthermore, if the background image and the union location of where the target human object should appear are provided by users, Interact-Custom also provides the optional functionality to specify them, offering high content controllability. Extensive experiments on our tailored metrics for CHOI task demonstrate the effectiveness of our approach. Our code is available at \url{https://github.com/XZPKU/Inter-custom.git}
\end{abstract}

\author{Zhu Xu}
\affiliation{%
  \institution{Wangxuan Institute of Computer\\ Technology, Peking University}
  \city{Beijing}
  \country{China}}
\email{xuzhu@stu.pku.edu.cn}

\author{Zhaowen Wang}
\affiliation{%
  \institution{Adobe Research}
  \city{San Jose}
  \country{US}}
\email{ zhawang@adobe.com}

\author{Yuxin Peng}
\affiliation{%
  \institution{Wangxuan Institute of Computer\\ Technology, Peking University}
  \city{Beijing}
  \country{China}}
\email{pengyuxin@pku.edu.cn}

\author{Yang Liu}
\authornote{Corresponding author}
\affiliation{%
  \institution{Wangxuan Institute of Computer\\ Technology, Peking University}
  \city{Beijing}
  \country{China}}
\email{yangliu@pku.edu.cn}

\renewcommand{\shortauthors}{Zhu Xu, Zhaowen Wang,  Yuxin Peng, \& Yang Liu}



\maketitle
\vspace{-4mm}
\section{Introduction}
Recently, subject customization image generation ~\cite{dreambooth,textinversion} have demonstrated remarkable quality in generating realistic images for given target concepts. On the basis of them, compositional image customization ~\cite{zhu2024multiboothgeneratingconceptsimage,fastcomposer, disenbooth}, which aims to generate images with multiple target concepts simultaneously, has aroused increasing attention for its facilitation in more broad user-specific applications like advertising production. 
\begin{figure}
    \centering
    \includegraphics[width=0.95\linewidth]{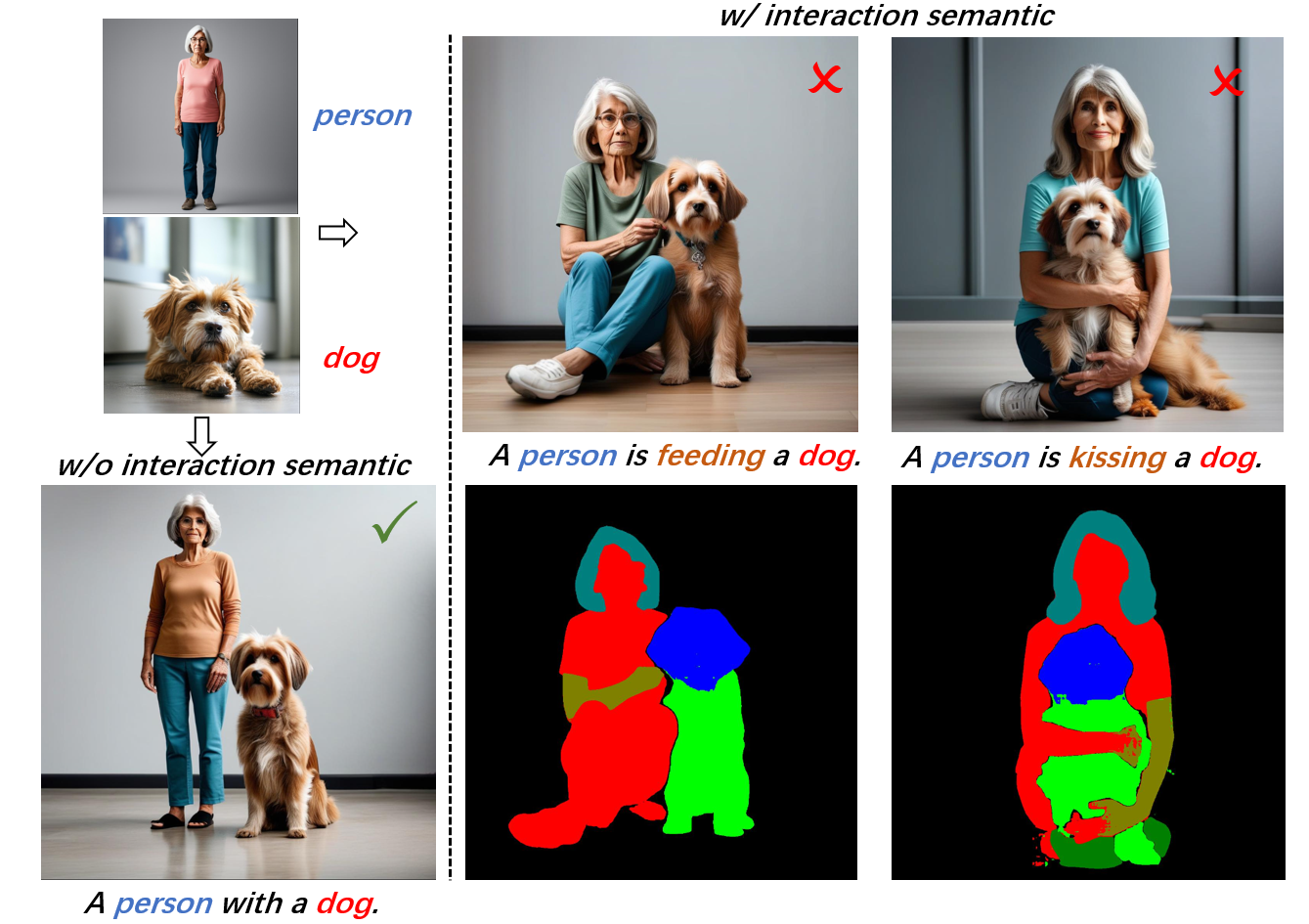}
    \caption{\textbf{Visualizations of interaction-involved generation of
    compositional customization approach MIP-Adapter\cite{ma2024subjectdiffusionopendomainpersonalizedtexttoimage}.} It fails to follow interaction semantics within text prompts. Specifically, wrong spatial configuration between \textit{human hand} and \textit{dog mouth} reveals that degraded interaction boundary impairs interactive semantic representation.}
    \label{fig:teaser}
\end{figure}
Though a great success in existing compositional customization image generation methods, when encountering real-world scenarios involving complex interaction among target entities, current approaches still struggle to generate satisfying contents, as they neglect to achieve precise control over fine-grained inter-entity interaction semantics. As shown in Fig.~\ref{fig:teaser}, when the text prompts involve interaction between the target human and dog, the recent advanced customization approach MIP-Adapter\cite{huang2024resolvingmulticonditionconfusionfinetuningfree} fails to convey corresponding action in generated images, causing a semantic deviation from text prompts.
To facilitate the model of such interaction control ability, we focus on scenarios that are human-centric while coupled with interaction with diverse objects for study, and propose the new task of Customized Human Object Interaction Image Generation(CHOI). CHOI simultaneously requires identity preservation for target human and object, as well as interaction semantic control among them. The most relevant task to CHOI is Virtual Try-on\cite{cordier20012d}, which aims to synthesize a naturally dressed person with target clothing and human appearance. However, Virtual Try-on is physically constrained by the form of interaction and can only achieve the fitting of a single object, i.e., cloth, to the human body, without supporting more diverse combinations and interactions forms between human and object. In contrast, our CHOI task is more free-form, encompassing the generation ability for a wide range of object and interaction categories.

To systematically study the CHOI task, we face following two primary challenges:
(1) \textbf{Lack of appropriate data}: CHOI requires maintaining identity features for target human and object, while controlling their poses to allow for flexible and diverse variations to meet the requirements of diverse interactions. However, existing Human-Object Interaction image datasets only provide static interaction content, making it difficult to separate self-contained target identity features from pose-oriented features that undergo flexible changes for interaction semantic expression. Such static data hampers the model’s ability to learn the feature separation.
(2) \textbf{Inappropriate spatial configuration between humans and objects hinders interaction semantic expression}: Apart from the feature separation need for target human and object themselves, modeling a reasonable spatial configuration between human and object is also a crucial factor for interaction semantic expression. For instance, in the \textit{human feeding dog} scenario shown in Fig.~\ref{fig:teaser}, the \textit{feeding} action requires not only that the distance between \textit{human} and \textit{dog} is not too far, but also the \textit{hand of human} and \textit{mouth of dog} are in close proximity. Therefore, a reasonable and fine-grained spatial configuration (specifically detailing how each body part should interact with the object) should be modeled to accurately control the expression of interaction semantics.
Furthermore, benchmarking comprehensive and systematic evaluation for CHOI is required since we are the first work to investigate into this task.

To address above issues, the following solutions are proposed:
Firstly, to tackle the problem of appropriate data shortage, we collect and process a large-scale dataset encompassing data samples tailored for CHOI task. We use specific pipelines to construct samples that maintain the identity consistency for target human and object, while enabling different poses. This allows the explicit separation of self-contained identity features from the pose-oriented interaction features for target human and objects, facilitating the model learning of such feature separation. We construct the corresponding data from image datasets and videos containing interaction semantics, then processed with tailored pipelines, ultimately obtaining around 1M data samples covering diverse interaction action categories, object types, and interaction poses. Then, to ensure a high-quality spatial configuration between human and object for interaction semantic expression, we introduce a two-stage model Interact-Custom, which tackles CHOI in a divide-and-conquer manner. In the first stage, we employ a diffusion model to generate the mask of target human object with interaction text prompts as control signals, guiding the spatial configuration to accurately convey target interaction behaviors. Then in the second stage, we aim to generate the target human and object interacting with their identities preserved. Extracted identity features ensure the identities are maintained, while for interaction semantic control, we utilize the mask as spatial configuration guidance, guiding the target human and object to follow such pattern while combining the inherent characteristics of the target human object to express corresponding interactive action. 
What's more, Interact-Custom provides the optional function to specify background content as well as the union location of human object interaction. With additional input of one background image and a bounding box format location, Inteact-Custom can yield results with the target subject object interacting within the specified region, while seamlessly integrating into the background content, offering high content controllability.
Finally, we tailored diverse metrics to systematically evaluate the generation quality of CHOI task.

In summary, our contributions are fourfold. (1) We introduce a novel task Customized Human Object Interaction Image Generation(CHOI), enhancing the interaction control ability of compositional customized image generation.
(2) we collect and process a large-scale dataset, encompassing data samples that own the same pair of human object involving different poses, to facilitate the study for CHOI task.
(3) We propose a two-stage model Interact-Custom that explicitly models the spatial configuration between human object to ensure the interaction semantic control. It also enables the optional function of background content and human object union location specification.
(4) Through comprehensive evaluation on our proposed metrics, the effectiveness of our approach is validated.

\vspace{-5pt}
\section{Related Work}

\noindent\textbf{Single Target Concept Customization} aims to 
customize the identity for one given target concept. 
DisenBooth\cite{disenbooth} generates images for specific concept by optimizing corresponding concept embedding. ControlCom\cite{zhang2023controlcom} proposes to extract target ID feature for identity preservation.  
AnyDoor\cite{chen2024anydoorzeroshotobjectlevelimage} and IMPRINT\cite{song2024imprintgenerativeobjectcompositing} further introduce detailed maps of target objects to preserve fine-grained features. However, such single concept customization task overlooks the disentanglement of identity features and their pose features, typically resulting in identical object poses in the generated image and reference image. Compared to it, our CHOI task not only disentangles the identity feature and pose feature to enable generation with flexible and diverse interaction poses, but also controls the interaction semantic between the target human and object, showing higher controllability in generated content. 

\noindent\textbf{Compositional Customized Image Generation} aims to generate images for multiple target concepts by providing one or a few exampler images for each target. 
FreeCustom\cite{ding2024freecustomtuningfreecustomizedimage} propose to inject target concept features in a tuning-free manner. Mix-of-show\cite{Mix-of-Show} utilizes decentralized LoRA to enable multiple concepts customization. Perfusion\cite{tewel2024keylockedrankeditingtexttoimage} optimizes text tokens in cross-attention to enable multiple concepts identity preservation. MIP-Adapter\cite{huang2024resolvingmulticonditionconfusionfinetuningfree} propose to resolve the confusion between different concepts.
Though these methods are applicable to human object interaction scenarios, when encountering more complex, deformable non-rigid interactions, the interaction semantics in the generated images may fail to be expressed due to inappropriate spatial configurations. Therefore, our Interact-Custom explicitly 
provides masks to serve as spatial configuration guidance, yielding accurate semantic expression and generating images encompassing diverse yet vivid interaction between human and object.

\noindent\textbf{Human-centric Customized Image Generation} aims to maintain specific identity characteristic for given target human in generated images. FastComposer\cite{fastcomposer}, Subject-Diffusion\cite{ma2024subjectdiffusionopendomainpersonalizedtexttoimage},Multi-Booth\cite{zhu2024multiboothgeneratingconceptsimage} finetunes the pre-trained diffusion model to incorporate ID-preserved features for target humans. 
More works\cite{li2023photomakercustomizingrealistichuman,wang2024instantidzeroshotidentitypreservinggeneration,gao2025conmo,lei2024exploring,libalancing,Qin2025ApplyHT,10410600,9996117,zheng2024TFTG,Lei_2024_CVPR} like Photomaker\cite{li2023photomakercustomizingrealistichuman}, InstantID\cite{wang2024instantidzeroshotidentitypreservinggeneration} propose to preserve target human ID in a tuning-free manner in consideration of efficiency. 
Though great successes, the identity preservation in these methods only concentrates on the human face, ignoring other crucial features that can identify the characteristics of the target human like their clothes and shoes, showing limited application in the face domain. Virtual Try-on\cite{cordier20012d} further requires the customization of clothes, but the interaction form of clothes and humans is restricted without generalization. Compared to them, our CHOI task not only enables human identity preservation in a more comprehensive manner, i.e., including their identities of other body parts as well, but also enables the object identity preservation and the diverse interaction expression between human and object, showing higher controllability over broader scope.

\noindent\textbf{Human Object Interaction Image Generation} aims to generate images featuring human-object interaction scenes following text prompts or additional control signals.SA-HOI\cite{xusemantic} proposes the first HOI image generation dataset with tailored evaluation protocols, and proposes to refine the human pose as well as interacting semantic in a training-free manner.

Interact-Diffusion\cite{Hoe_2024_CVPR} proposes to incorporate bounding-box information to control the specific locations of interacting human-object. Specifically, their model is incapable of understanding the fine-grained spatial configuration for expressing interaction semantics, as the individual bounding boxes for humans and objects are all provided as their input conditions, which lowers their task difficulty in comparison with only adopting one box as the union region condition. 
What's more, these methods are incapable of preserving the ID of interacting humans and objects, and our approach proposes to further enable model of such subject customization ability, broadening application scope.

\section{Task Definition}
Given the target human image $I_{h}$ and object image $I_{o}$, the text prompt $T_{inter}$ describing the interaction action between human and object, we aim to generate image $I_{gen}$ depicting the target human interacting with target object by the action of $T_{inter}$, while faithfully preserving the identity of $I_h$ and $I_o$. Optionally, we can include additional input of background image $I_{bg}$ paired with bounding box positions $B$ to indicate the union region of the human-object interaction, and require the target human and object to interact within the union box specified by $B$, while seamlessly integrating into the background $I_{bg}$. 

\section{Dataset Construction}
For CHOI task learning, the simultaneous demand on identity preservation and interaction semantics control of the target human and object requires the model to decouple the self-contained identity features and pose-oriented interaction features for humans and objects. To facilitate learning of such capability, the training samples should own identical interactive human object pair with different poses. However, existing human-object interaction image datasets\cite{chao:iccv2015,gupta2015visual,E210455} fail to provide such samples as the pose for each human-object pair is static. To acquire such training samples, we delve into several attempts and ultimately collect a large-scale and diverse dataset for training. We separately introduce the collection pipeline as follows:

\subsection{Image-dataset Adaptation}
\begin{figure}[h]
    \centering
    \includegraphics[width=0.95\linewidth]{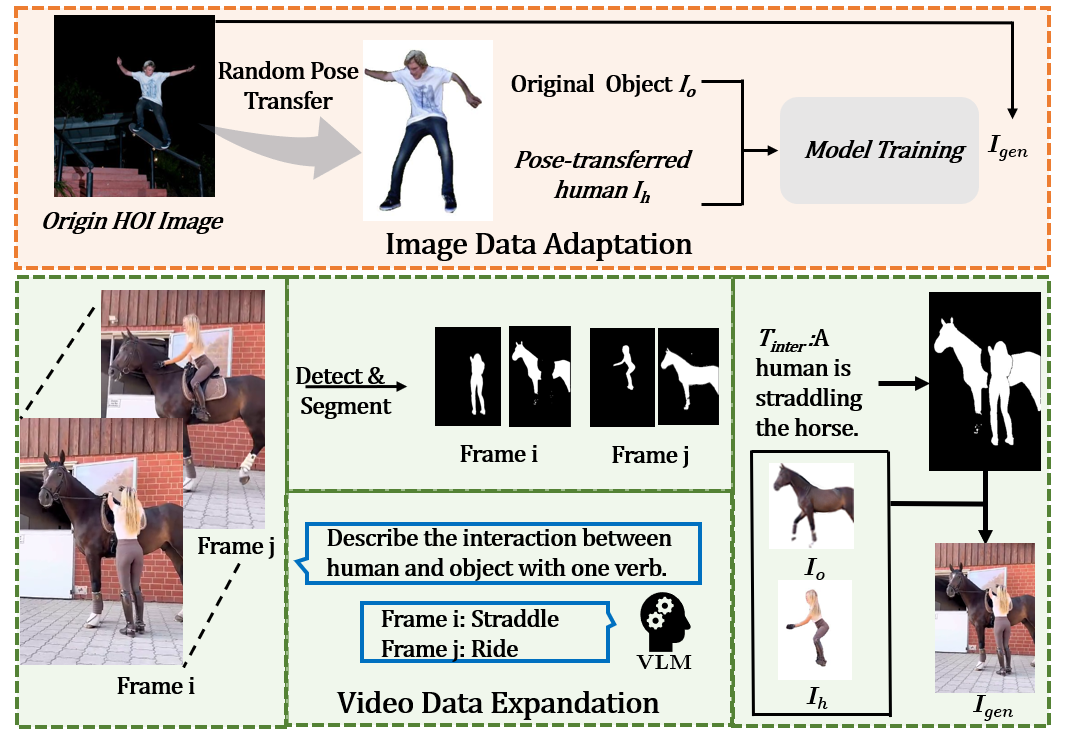}
    \caption{\textbf{Processing procedure for image and video data.} For images, human poses are randomly changed to construct training samples. For videos, frames for identical interacting human-object pair are utilized to form training samples.
    }
    \label{fig:video_collect}
\end{figure}

Human object interaction image datasets provide a large amount of data with diverse interaction categories. However,  their static pose characteristics hinder their direct usage for the learning of identity feature and interaction feature separation. Therefore, for existing human object interaction image dataset HICO-Det\cite{chao:iccv2015}, we seek to modify the pose of either the human or the object to obtain reference images with different poses of the target human or object. This allows us to create data sample constructed by a human object interaction image, along with different posed reference image of human and object, which meets our feature separation requirements. Specifically, we adopt the segmentation tool\cite{segmentanything} to segment target human and object out, then adopt \cite{chen2024anydoorzeroshotobjectlevelimage} to change the pose of the target human and object by providing a category-identical while random-posed mask\footnote{we collect the human mask within HICO-Det as mask pool for random mask selection.}
as shape guidance.
However, the shape divergence within each object category harms the quality of pose-transferred objects, so we only adopt the pose-changed humans to form such training samples, which is shown in Fig.~\ref{fig:video_collect}. Ultimately, we yield one training sample with the pose-changed human and original object image as the model input, targeting to generate the original HOI image. This portion of data not only facilitates model's learning of human-centric disentangled identity features and interaction semantics, but also provides a wealth of samples modeling spatial configurations across a broader range of interaction categories.

\subsection{Video-dataset Expansion}
We further propose to incorporate more challenging data, which encompass pose changes in both humans and objects with higher diversity. Considering that videos naturally own such pose dynamics, we collect such training samples by including videos from the website. We select videos with salient human-object interaction scenarios as the foreground, and crop the clip where the identical human-object pair is interacting without camera change, to generate our training samples. The data process procedure is shown in Fig.~\ref{fig:video_collect}. For each video clip, we randomly select two frames $f_{i}$ and $f_{j}$, utilizing Grounded-SAM\cite{ren2024grounded} to acquire corresponding human mask $mh_{i/j}$and object mask $mo_{i/j}$. To annotate the exact interaction between humans and objects, we utilize Intern-VL\cite{chen2023internvl} to distinguish the action verb for each frame $v_{i/j}$, then form the interaction prompt $T_{inter-i/j}$ by composing them into a template ``A person is \textit{verbing} the \textit{object}".
To this end, constructed data samples can be utilized for two types of training: for the learning of spatial configuration between humans and objects, data samples encompass segmented foreground masks of human object pair in each frame as ground truth, and interaction prompts serve as input condition; for feature disentangling learning, any two frames of images within the same video clip can be used to construct samples with the same identity but different poses: by serving segmented human and object in $v_i$ as $I_{h}$ and $I_o$, we incorporate the $T_{inter-j}$ as interaction prompts and $v_j$ as the ground truth image of $I_{gen}$. Owing to the natural and seamless pose variations that occur for the same human object pair in video data, this portion of data facilitates the model in concurrent feature disentangle learning for both human and object.

\subsection{Dataset Summary}
In summary, by adopting both the processed data from images and videos, we yield a large-scale dataset that owns samples of identical human object pair showing different poses. As shown in Tab.~\ref{dataset_compare}, it contains 1M samples in total, covering 630 interaction categories with 85 object categories in daily life, featuring a comprehensive yet diverse collection of object, interactive action as well as action-object combinations, which surpasses all the previous HOI datasets. Such thorough coverage effectively captures common interactions encountered in daily life, and the specific ``identical human object with different poses" characteristic makes it effective for CHOI task learning. 

\begin{table}[]
\caption{\textbf{Comparison with existing HOI datasets.} ``Scale" denotes data sample number, ``Combination" denotes the number of ``action-object" categories, ``support CHOI" denotes whether the data is suitable for CHOI training.}
\resizebox{0.47\textwidth}{!}{
\begin{tabular}{cccccc}
\hline
         
         & Scale & Object & Interaction &Combination &Support CHOI \\
         \cline{1-6}
  V-COCO~\cite{gupta2015visual} & 10k & 80 & 29 & - & \ding{55}\\
  HICO-Det~\cite{chao:iccv2015} & 48k & 80 & 117 & 600 & \ding{55}\\
  Action Genome~\cite{ji2019actiongenomeactionscomposition} & 234k & 35 & 25 & 157 & \ding{55} \\
  VidHOI~\cite{chiou2021st} & 7.3M &78 & 50 & 557 & \ding{55}\\
  Ours & 1M & 85 & 121 &630 &\ding{52} \\
\hline
\end{tabular}}
\label{dataset_compare}
\end{table}

\section{Method}
\begin{figure*}[t]
\centering 
\includegraphics[width=0.85\linewidth]{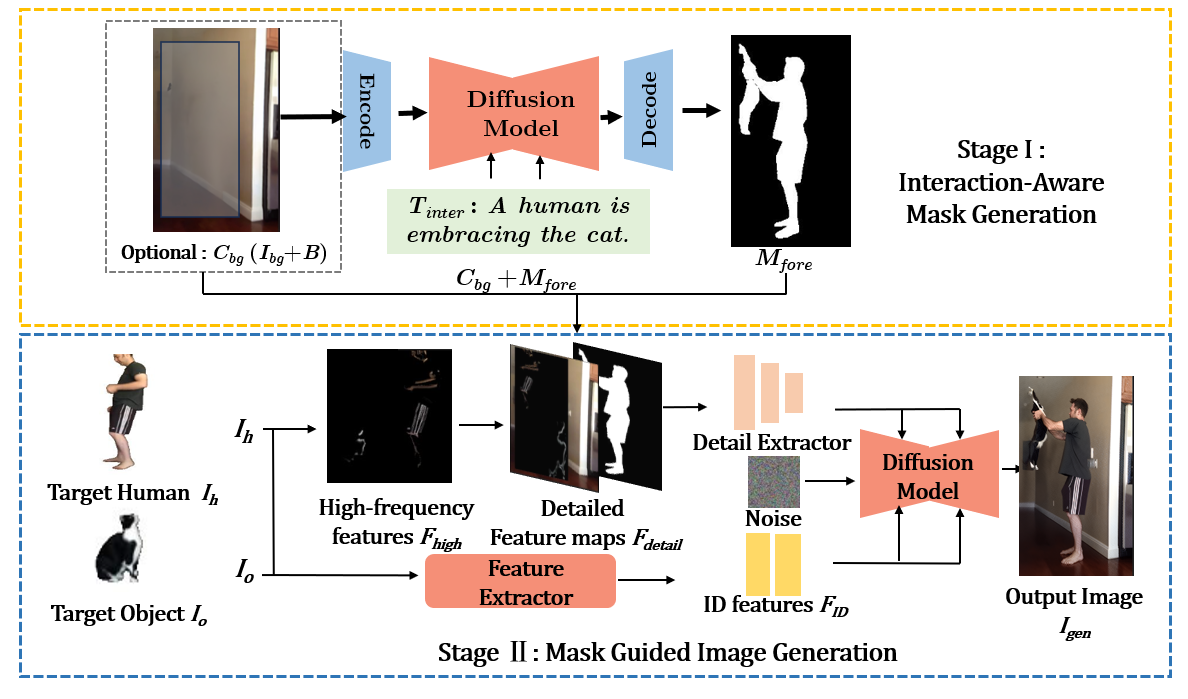} 
\caption{%
    \textbf{Overall pipeline of Interact-Custom}. In Interaction-Aware Mask Generation (IAMG) Stage, the Generation-Based model serves prompt $T_{inter}$ as the conditions, generating human-object mask $M_{fore}$ that accurately conveys target interaction semantic. $M_{fore}$ is adopted as spatial configuration guidance for image generation. In Mask Guided Image Generation (MGIG) stage, $I_{h}$ and $I_{o}$ are utilized to extract ID features and fine-grained detail features $F_{high}$, serving as disentangled self-contained identities features. $M_{fore}$ is then incorporated to guide to express interaction semantic. Optionally, background image $I_{bg}$ and location $B$ to specify the union region of the human-object interaction can serve as additional input to control the background and specific location.}
    
\label{fig:pipeline}
\end{figure*}

The pipeline of Interact-Custom is demonstrated in Fig.~\ref{fig:pipeline}. It consists of two phases, termed Interaction-Aware Mask Generation (IAMG) and Mask Guided Image Generation (MGIG). In IAMG, we aim to generate the mask for target human object as an explicit representation of spatial configuration to match the target interaction prompt $T_{inter}$. We employ a diffusion-based model to accomplish it, which serves $T_{inter}$ as control signal to specify the interaction semantic of the generated mask. 
Then in MGIG, we serve the IAMG-generated mask as guidance to synthesize the content of target human $I_h$ and object $I_{o}$ interacting in the output image $I_{gen}$. The self-contained identity features of $I_h$ and $I_o$ are extracted to preserve the appearance of target human and object, while the interaction semantic features are altered to follow tha pattern of IAMG-generated mask, subsequently leading the interaction semantic to follow $T_{inter}$. Additionally, if users have specific control requirements regarding the background and the precise union locations of the human object interactions, we offer optional functionality to accept additional background image $I_{bg}$ and bounding box format positional inputs $B$. In IAMG model, by equipping $I_{bg}$ with $B$ as extra control signals, the generated mask is deemed to locate the interaction within the given bounding box region. Then in MGIG, by further incorporating $I_{bg}$ and $B$, we can guide humans and objects to interact at designated locations while seamlessly integrating them into the provided background.

\subsection{Interaction-Aware Mask Generation}

\noindent\textbf{Basic Generation:} To predict a mask depicting the interaction within $T_{inter}$, we introduce a diffusion model  $\mathcal{D}_{\theta}$ to accomplish it. Specifically, we pass $T_{inter}$ through a frozen text encoder $\mathcal{E_{T}}$ as generation conditions. At each denoising timestep $t$, the predicted noise $\Tilde{\mathbf{\epsilon}}_{pred} (t)$ is calculated jointly on previous noise latent $\mathbf{z}_{t}$,  and input condition as follows,
\begin{align}
    \Tilde{\mathbf{\epsilon}}_{pred} (t) = \mathcal{U}_{\theta} (\mathbf{z}_{t},\mathcal{E_{T}}(T_{inter}), t),
\end{align}
where $\mathcal{U}_{\theta}$ represents the U-Net of the diffusion model $\mathcal{D}_{\theta}$. 
The diffusion model $\mathcal{D}_{\theta}$ is trained to predict the actual noise $\epsilon_{t}$ with following formulation,
\begin{align}
    \mathcal{L}_{t}(\theta, \phi) = \mathbf{E}_{t\sim[1,T], T_{inter}, \epsilon_t}[\Vert \mathbf{\epsilon}_t - \Tilde{\mathbf{\epsilon}}_{pred} (t)\Vert^2],
    \label{eq1}
\end{align}
where $T$ is total number of diffusion timesteps, $\epsilon_t \sim \mathcal{N}(0,I)$ is sampled from a normal distribution. 
$ \mathcal{U}_{\theta}$ can generate masks with diverse patterns for each interaction category, yielding satisfying results.

\noindent\textbf{Controllability Expandation:} 
Considering that interactions in different regions of the same scene can influence both the form and content of the interaction, understanding the background context around the interaction region can help the model generate masks that are better suited to the whole scene. To further enable the function of background content and interaction location specification, we incorporate $I_{bg}$ with $B$ as location condition $C_{bg}$ as extra condition, which masks the bounding box region of $B$ within $I_{bg}$, to guide the model not only generate the human and object mask locating within the specified region of $B$, but also utilize surrounding environment information from $I_{bg}$ to adaptatively alter the mask, ensuring that the human and object in the final image can naturally blend into the corresponding background. We pass $C_{bg}$ through a frozen image encoder $\mathcal{E_{I}}$ to obtain the encoded features $\mathcal{E_{I}}(C_{bg})$, and concatenate it on the noised latent features as the control signal, where the formulation is altered as follows:
\begin{align}
    \Tilde{\mathbf{\epsilon}}_{pred} (t) = \mathcal{U}_{\theta} (\mathbf{z}_{t},\mathcal{E_{I}}(C_{bg}),\mathcal{E_{T}}(T_{inter}), t),
\end{align}
and we train $\mathcal{D}_{\theta}$ with following formulation,
\begin{align}
    \mathcal{L}_{t}(\theta, \phi) = \mathbf{E}_{t\sim[1,T], C_{bg}, T_{inter}, \epsilon_t}[\Vert \mathbf{\epsilon}_t - \Tilde{\mathbf{\epsilon}}_{pred} (t)\Vert^2],
    \label{eq1}
\end{align}

Ultimately, the generated mask serves as $M_{fore}$, featuring the human-object pair interacting with the target interaction semantic. $M_{fore}$ are then utilized to guide the following human-object interaction image generation process with appropriate spatial configuration depicting target interaction semantics. 

\subsection{Mask Guided Image Generation}
\noindent\textbf{Preliminary:}
We adopt AnyDoor\cite{chen2024anydoorzeroshotobjectlevelimage} as the baseline of our MGIG model. The core idea of AnyDoor is to adopt different granularity features for identity preservation: image features extracted from DINOv2\cite{dinov2} encoder are adopted as coarse ID features, the high-frequency feature maps of target subject derived from Sobel\cite{sobel} kernel is incorporated as fine-grained detail complementary. However, it only supports identity preservation of single object, which is against our objectives. We make several modifications to this method to align it with our task requirements, as introduced below.

\noindent\textbf{Approach Adaptation:}
To enable the model handling target human and object simultaneously, we firstly adopt DINOv2\cite{dinov2} to separately extract the ID features for $I_{h}$ and $I_{o}$, acquiring $F_{ID} \in \mathbb{R}^{2\times N_p \times C}$, where $N_p$ is the patch number
and $C$ is feature dimension. 
Then we concatenate $I_{h}$ and $I_o$, passing it through Sobel kernel to acquire the high-frequency maps of both the target human and object, denoted as $F_{high}$, to serve as details complementary for the ID features $F_{ID}$. $F_{ID}$ and $F_{high}$ together serve as the self-contained identity features for $I_h$ and $I_o$ for target human object identity preservation. For interaction semantic control, we utilize generated mask $M_{fore}$ as shape guidance during generation, which not only ensures the pose of human or object to follow its pattern, transferring them into different poses suited for the target interaction category, but also guide their spatial configuration to be appropriate for interaction semantic expression. Specifically, $F_{high}$ and $M_{fore}$ are channel-wise concatenated to serve as control signal over both shape details and appearance details, denoted as $F_{detail}$, and then injected through a U-Net-shaped feature extractor to generate detailed feature maps of different resolutions, and identity features $F_{ID}$ are injected through cross-attention layers, which is formulated as follows,
\begin{align}
    \Tilde{\mathbf{\epsilon}}_{hoi} (t) = \mathcal{U}_{hoi} (\mathbf{z}_{t},F_{detail},F_{ID}, t),
\end{align}
where $\mathcal{U}_{hoi}$ is the U-Net for MGIG. Multi-granularity ID features ensures the identity preservation for tagret human object, while $M_{fore}$ guarantees a reasonable spatial configuration for interaction semantic expression.

\noindent\textbf{Controllability Discussion:} 
With additional input of $I_{bg}$ and $B$, MGIG is further enabled of background and location specification ability. We accomplish it by enhancing $F_{detail}$ with $I_{bg}$ and $B$. Specifically, we reshape $F_{high}$ into size of bounding box $B$, and add it on the $B$ region of $I_{bg}$, which formulated as,  
\begin{equation}
    F_{detail} = Concat (M_{fore}, (Reshaped(F_{high})\oplus I_{bg}))
\end{equation}
We enable optional background and location specification ability by randomly dropping out $I_{bg}$ and $B$ during training.

\noindent\textbf{Training Procedure of MGIG:}
Given that the shape of the mask $M_{fore}$ may occasionally be inconsistent with the shape of the target human and object, leading to low-quality or blurred regions around the boundary in the generated human object. We address it by decomposing the training for MGIG into two stages. In the first stage, we train MGIG with the ground truth mask $M_{fore}$ acquired from the training samples, ensuring that the generated image adhere faithfully to the shape constraints provided by $M_{fore}$. Then in second stage, we leverage the mask generated by our IAMG as $M_{fore}$ for training, which allows for a trade-off between the accuracy of mask control and the generation quality, adaptively teaching model to solve the potential conflicts.

\section{Experiments} \label{sec:exp}

\subsection{Implementation Details}
For IAMG, we select Stable-Diffusion-v2.0\cite{ldm} as our base model. For MGIG, we choose AnyDoor~\cite{chen2024anydoorzeroshotobjectlevelimage} as our base model, and we train the models from the corresponding checkpoint on our proposed dataset. We adopt AdamW~\cite{adam} as our optimizer with the learning rate of 1e-5. IAMG is trained for 800k steps, and MGIG is trained for 600k steps.

\noindent\textbf{Tasks and Evaluation Metrics:} 
For \textit{\textbf{subject customization}}, we adopt CLIP-Score and DINO-Score to compare the generated images with original images, including the similarity of segmented target human, object as well as their combinations. 
For \textit{\textbf{interaction semantic control}}, we adopt three tasks for comprehensive evaluation.
(1) \textit{Spatial-sensitive semantic}: we utilize a human-object interaction detector RLIP-v2\cite{Yuan2023RLIPv2} to detect all possible HOI instances within the generated images to infer the expressiveness of interaction semantics, and we adopt mAP as the evaluation metric following previous work\cite{chao:iccv2015}.
(2) \textit{Holistic Semantic:} We adopt LLaVA\cite{liu2023visualinstructiontuning} to judge whether the generated image contains target interaction semantics, and we calculate the accuracy for evaluation. 
(3) \textit{Spatial-configuration alignment:} 
To evaluate the quality of our Interaction-Aware Mask $M_{fore}$, we calculate the distribution distance between our masks and ground truth masks.
Further, we conduct a user study to evaluate the quality of our generated images from different perspectives. Finally, we provide some ablation studies and qualitative results. 

\begin{table}[]
\caption{\textbf{Quantitative evaluations on subject customization.} We compare the identity preservation ability with existing compositional subject customization approaches.}
\resizebox{0.47\textwidth}{!}{
\begin{tabular}{ccccc}
\hline
         
          \cline{1-5} 
          &  CLIP-Score \( \uparrow \) & \multicolumn{3}{c}{DINO-Score \( \uparrow \)  }\\
          \cline{1-5}
           & & Human & Object & Pair\\
          \cline{1-5}
Fastcomposer\cite{fastcomposer}   & 74.19  & 58.23 & 64.60         & 61.28                    \\
\cline{1-5}
MIP-Adapter\cite{ma2024subjectdiffusionopendomainpersonalizedtexttoimage}    & 78.43  & 71.23 & 67.19         & 65.60              \\
\cline{1-5}
AnyDoor\cite{chen2024anydoorzeroshotobjectlevelimage}   & 82.31  & 70.08 & 72.27          & 74.14                    \\
\cline{1-5}
Ours    & \textbf{87.60}  & \textbf{78.90} & \textbf{81.39}          & \textbf{83.27}       \\ 
\cline{1-5}
\hline
\end{tabular}}
\label{subject}
\end{table}

\subsection{Comparisons with Compositional Customized Approaches on Subjects Fidelity}
\label{exp_subject}
Compositional customization approaches can be divided into tuning-based and reference-based approaches. Tuning-based approaches require extra fine-tuning for each given target concept pair, which hinders the large-scale evaluation. Thus, we only compare open-sourced reference-based approaches \cite{fastcomposer, chen2024anydoorzeroshotobjectlevelimage, ma2024subjectdiffusionopendomainpersonalizedtexttoimage} quantitatively. The results are shown in Tab.~\ref{subject}, and we can conclude that: (1) For CLIP-Score, we harvest the highest performance, indicating that our generated image is most visually similar to ground truth images. 
(2) For DINO-Score, we utilize the mask of the target human and object to segment the corresponding region for similarity calculation. With our large-scale dataset training, our model successfully learns how to preserve the identity of diverse human-object combinations, thus harvesting the best performance for DINO-score of human, object as well as their combinations, outperforming our baseline model AnyDoor\cite{chen2024anydoorzeroshotobjectlevelimage} by 8.82\%/9.12\%/9.13\%.

\begin{table}[t]
\centering
\caption{\textbf{Performance on interaction semantic control.} \textit{Spatial-Sensitive semantic} utilizes FGAHOI\cite{Ma2023FGAHOI} as detector and is measured with mean average precision (mAP). \textit{Holistic Semantic} utilizes LLaVA\cite{liu2023visualinstructiontuning} and is measured with accuracy.}
\setlength{\tabcolsep}{3pt} 
\resizebox{0.45\textwidth}{!}{
\begin{tabular}{l|c|ccc}
\hline
\multirow{2}{*}{Model} & \multirow{2}{*}{HolisticSemantic (\%)} & \multicolumn{3}{|c}{Spatial-Sensitive Semantic}                   \\ \cline{3-5} 
                       & &Full           & Rare           & Non-Rare                \\ \hline

\textit{Customization approach} & \\
\cline{1-5}
FastComposer & 4.9 & 0.87& 0& 1.23 \\
MIP-Adapter & 81.73 & 1.10& 0.65& 1.31\\
AnyDoor& 82.04& 17.54& 10.63& 19.18 \\
\cline{1-5}
\textit{Interaction Control approach} & \\
\cline{1-5}
GLIGEN  &  83.17  & 19.48 & 10.94   & 20.63  \\
InteractDiffusion &  84.18   & \underline{29.47} & \underline{23.13} & \underline{30.47}           \\ 
\cline{1-5}
Ours &  \underline{86.02}& 22.07& 11.87& 23.87 \\
\hline
HICO-DET (Ground Truth Image)               & \textbf{95.92}                            & \textbf{30.17}        & \textbf{22.23}          & \textbf{31.62}                          \\ \hline
\end{tabular}}

\label{tab:hoid}
\end{table}
\begin{table}[hhh]
\caption{%
    \textbf{Results of \textit{Spatial-configuration Alignment.} }We adopt K-L divergence for calculation. ``Random" represents randomly select mask. ``IAMG-R" and ``IAMG-G" represents retrieval-based and generation-based IAMG model.}
\centering\footnotesize
\setlength{\tabcolsep}{5.5pt}
\resizebox{0.49\textwidth}{!}{
\begin{tabular}{lcccccc}
\toprule
&Random & FastComposer&MIP-Adapter&AnyDoor&IAMG-R& IAMG-G  \\
\midrule

Distance($(10^{-2})\downarrow$   & 37.02& 54.60&18.63 & 12.56&  8.97   &  6.43    \\
\bottomrule
\end{tabular}}

\label{tab:iaf}
\end{table}
\subsection{Comparisons with Other Approaches for Interaction Control Ability} 
\label{exp_interaction}
We adopt \cite{ldm,li2023gligen,Hoe_2024_CVPR,fastcomposer,zhu2024multiboothgeneratingconceptsimage,chen2024anydoorzeroshotobjectlevelimage} for comparison, which comprises two lines of works: \textit{Customization Approaches} aim to preserve target identity, while \textit{Interaction Control Approach} targets for interaction semantic control. The HICO-DET\cite{chao:iccv2015} subset of our test-set is adopted for evaluation as they contains ground truth annotation for HOI instances.
For \textit{Holistic Semantic} task, we judge whether each generated image contains the target human-object interaction semantic. The results are shown in Tab.~\ref{tab:hoid}. Our approach (Line 6) harvests the highest accuracy compared with other approaches, and is strikingly close to the performance of real images (Line 8), indicating our efficacy for interaction semantic control. 
Then for \textit{Spatial-Sensitive Semantic} task, which further takes the spatial configuration of interacting humans and objects into consideration, we adopt FGAHOI\cite{Ma2023FGAHOI} as the detector for evaluation. The results are shown in Tab.~\ref{tab:hoid}, from which we can conclude that (1) we outperform all the \textit{Customization approaches}, such as 4.53\%/1.24\%/4.69\% improvement in comparison with AnyDoor\cite{chen2024anydoorzeroshotobjectlevelimage}, indicating that we successfully equip customization approaches with interaction semantic control ability. (2) For \textit{Interaction Control Approach}, we outperform GLIGEN\cite{li2023gligen} but show inferior performance compared to InteractDiffusion\cite{Hoe_2024_CVPR}, which is the SOTA Interaction Control model. Since InteractDiffusion adopts the separate ground-truth bounding boxes of human and object for location control, it consequently owns lower difficulty for HOID than us and cannot be fairly compared. (3) There still exists a performance gap between our generated images and the real images, requiring further development.

\begin{figure*}[!ht]
    \centering
    \includegraphics[width=0.85\linewidth]{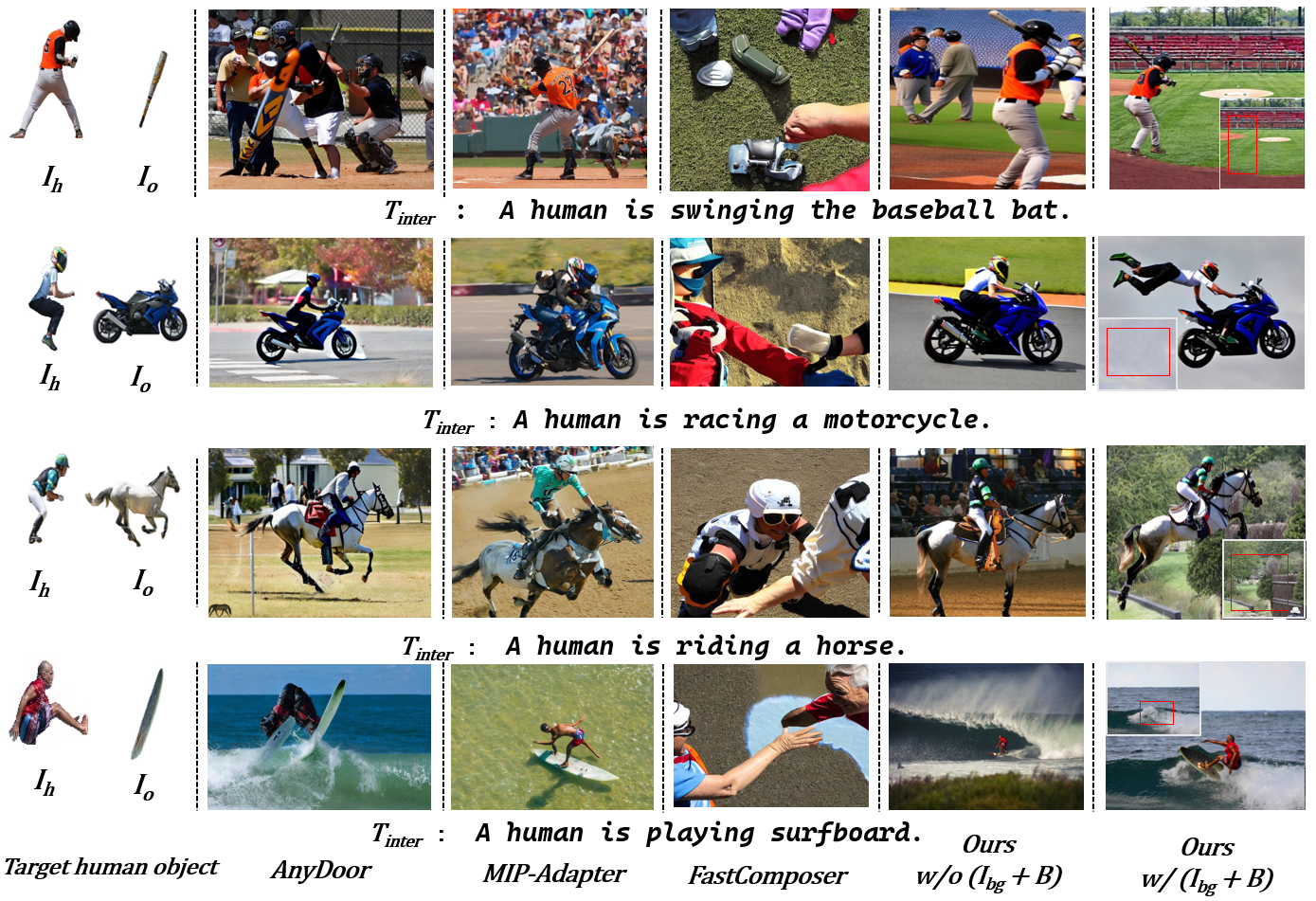}
\vspace{-4mm}
    \caption{\textbf{Qualitative comparison of different approaches.} Our approach yields better identity preservation ability as well as interaction semantic control. With additional ground-truth $I_{bg}$ and $B$, we can generate target human interacting object at the specified location (marked by red box), while seamlessly integrating into the given background.}
    \label{fig:compare}
\end{figure*}

Further, \textit{Spatial-configuration Alignment} task is conducted to measure the quality of spatial configurations of our IAMG phase. Specifically, to incorporate more personalized user requirements for obtaining preferred human object interaction masks, we also introduced a retrieval-based option in IAMG, which allows users to input free-form text to search for suitable masks in a pre-processed mask bank, we separately denote original generated-based IAMG and retrieval-based IAMG as IAMG-G and IAMG-R. We calculate the K-L divergence between generated masks and ground-truth masks as distribution distance, whose results are shown in Tab.~\ref{tab:iaf}. We compare with the human-object masks segmented from other approaches' generated images, and we also implement random mask selection from HOI mask bank (``Random") as a comparison. We can conclude that (1) In comparison with the masks segmented from other approaches' results, we yield better alignment with ground-truth masks, indicating that our IAMG encompasses better interaction semantic expression ability. (2) In comparison with ``random selection", our retrieval-based model can outperform random selection by 28.05\%, which benefits from that our query is equipped with detailed description of the target human-object pair as well as interaction semantics, which can provide accurate guidance to retrieve the appropriate masks. (3) Generation-based model outperforms retrieval-based model by 2.54\%. As the retrieval model can only retrieve the desired mask by calculating text similarity between the user-specified query and mask descriptions, which might fail to retrieve the mask with complete morphological correspondence, thus the performance is inferior to the generation-based model. Though \textit{Spatial-Semantic Alignment} can assess the quality of spatial configurations in a distributional sense, it still lacks more granular and precise evaluation, which remains as future work to expand upon.
\begin{table}[hhh]
\caption{%
    \textbf{User study with other compositional customized approaches.} 
    Each metric is rated from 1 (worst) to 5 (best).
}
\centering\footnotesize
\setlength{\tabcolsep}{5.5pt}
\resizebox{0.46\textwidth}{!}{
\begin{tabular}{lccccc}
\toprule
& Human~($\uparrow$) & Object~($\uparrow$) & BG~($\uparrow$) & Interaction~($\uparrow$) &  Overall~($\uparrow$) \\
\midrule
Fastcomposer\cite{fastcomposer}& 2.33& 1.28& 1.08& 1.58& 2.03\\
MIP-Adapter\cite{ma2024subjectdiffusionopendomainpersonalizedtexttoimage}& 3.72& 2.83& 1.28& 3.68& 3.03 \\
AnyDoor~\cite{chen2024anydoorzeroshotobjectlevelimage}          & 3.55& 3.23  & 3.78 & 3.58&3.40\\

 Ours   & \textbf{3.78} & \textbf{3.95} &  \textbf{4.05}  & \textbf{4.13}&\textbf{3.80} \\
\bottomrule
\end{tabular}}

\label{tab:userstudy}
\end{table}

\subsection{User Study }
\label{exp_user}
 We carry out a user study for a subjective evaluation. We invited 5 participants to rate the quality of the provided images through a questionnaire. 8 samples of different HOI categories were randomly selected, and images were generated accordingly using different models. Participants were instructed to rate these images on a scale from 1(worst) to 5(best). Evaluation criteria included (1) Human and Object appearance, which assesses the ID preservation of the target human and object. (2) the Background quality, which assesses how the background content is preserved during the generation process. (3) the Interaction Semantics, we judge the semantic relevance between the generated images and the given textual prompts. (4) overall quality through all the above aspects. With a total of 0.8K responses, insights into the performance comparison among the different models were obtained. Table.~\ref{tab:userstudy} shows the performance. We can conclude: (1) Our approach outperforms other methods in subjective evaluation across all metrics, which shows consistent performance trends with previous quantitative evaluations; (2) Our approach yields subject customization, interaction semantic control, and background control ability, suggesting its reliability.


\subsection{Ablation Study}
\label{exp_ablate}
We investigate how the performance of our approach is affected by different data sources.
The data sources contain two parts, non-interaction-aware data (NIA) and interaction-aware data (IA). The NIA is the pre-trained data for our baseline \cite{chen2024anydoorzeroshotobjectlevelimage}, which focuses on single instances and lacks interaction semantics. The IA data, i.e., our collected dataset, contains image data as well as video data.  We ablate on the different training data sources, the performance change is shown in  Tab.~\ref{tab:datasource}. The conclusions are (1) The large-scale NIA data (Line 1) enables the model of object-level customization ability, thus lowering the learning difficulty of human-object compositional ability: By combining NIA with our IA image data (Line 3), we can harvest performance gain compared to solely NIA (6.23\%/8.78\%/7.90\%/3.69\%) (line 1) or IA (3.75\%/5.04\%/3.81\%/2.59\%) (Line 2). (2) Since the HOI categories within image data of IA is more diverse than video data, training with Image data (Line 3) outperforms Video data (Line 4) by 3.57\%/4.97\%/4.95\%/0.02\%. (3) Considering that Video data encompasses higher pose divergence, further incorporating such pose-diversed Video data (Line 5) improves performance by 1.73\%/0.34\%/1.23\%/0.29\% compared with Image data (Line 3), harvesting performance gain in both aspects of subject customization as well as interaction semantic expression.

\begin{table}[hhh]
\caption{%
    \textbf{Ablation on training data sources.} ``NIA" represents Non-Interaction-Aware data, ``IA" represents Interaction-Aware data, including Image and Video.
}
\centering\footnotesize
\setlength{\tabcolsep}{5.5pt}
\resizebox{0.47\textwidth}{!}{
\begin{tabular}{l|cccc}
\toprule
& Human~($\uparrow$) & Object~($\uparrow$) & Pair~($\uparrow$) & VLM judgement (\%)$\uparrow$  \\
\midrule
Only NIA& 70.80 &72.27 & 74.14& 82.04\\
\cline{2-5}
Only IA& 73.42& 76.01& 78.23& 83.14\\
\cline{2-5}
NIA + IA (Image)&   77.17    & 81.05      &   82.04 & 85.73 \\
\cline{2-5}
NIA + IA (Video)    &  73.60  &  76.08    &  77.09    &   85.75  \\
\cline{2-5}
NIA + IA (Image + Video)      &   \textbf{78.90}    &  \textbf{81.39}   & \textbf{83.27} & \textbf{86.02}\\
\bottomrule
\end{tabular}}

\label{tab:datasource}
\end{table}

\subsection{Qualitative Results}
\label{exp_quality}
The qualitative comparisons are presented in Fig.~\ref{fig:compare}. We adopt the target human, object image $I_h, I_o$(Column 1) and interaction prompt $T_{inter}$ for generation. Previous approaches fail to simultaneously handling the target identity preservation and interaction semantic control, leading generated content unsatisfying. In contrast, our approach can precisely depict the human object interaction and preserve their identities. Further, by incorporating of background image $I_{bg}$ and specified location $B$ (Column 6), the human object interaction can be generated at the specified location and seamlessly integrated into the background.

\section{Conclusion}
We introduce the novel task of customized human object interaction image generation (CHOI), which both requires identity preservation for human object and the interaction semantic control between them. To tackle it, we first provide a large-scale and diverse dataset tailored for the training of CHOI task, which encompasses data samples of identical human object pair with different poses.
Then we propose an model Interact-Custom, which models the spatial configuration of the target human object to provide interaction semantic guidance, consequently enhancing the interaction semantic expression. Experiments on our proposed metrics validate the effectiveness of our method.

\begin{acks}
This work was supported by the
grants from the National Natural Science Foundation of
China (62372014, 62525201, 62132001, 62432001), Beijing Nova Program and Beijing Natural Science Foundation (4252040, L247006).
\end{acks}
\bibliographystyle{ACM-Reference-Format}
\bibliography{main}

\end{document}